%% file: 2019_acl_paragraph_embeddings.tex
\documentclass[11pt,a4paper]{article}
\usepackage[hyperref]{style/acl2019}
\usepackage{times}
\usepackage{latexsym}
\usepackage{amsfonts}
\usepackage{amsmath}
\usepackage{url}

\usepackage{adjustbox}
\usepackage{graphicx}
\usepackage{multirow}
\usepackage{pgfplots}
\usepackage{pgfplotstable}
\usepackage{subcaption}
\usepackage{graphicx}
\usepackage{blindtext}
\usepackage{booktabs}
\usepgfplotslibrary{groupplots,colorbrewer}

\aclfinalcopy 
\def\aclpaperid{1853} 

\newcommand{\bvec}[1]{\boldsymbol{#1}}
\usepackage[shortlabels]{enumitem}
\setitemize{itemsep=2pt,topsep=2pt,parsep=0pt,partopsep=5pt}

\title{Encouraging Paragraph Embeddings to Remember\\ Sentence Identity Improves Classification}

\author{Tu Vu \and Mohit Iyyer \\
  College of Information and Computer Sciences  \\
  University of Massachusetts Amherst \\
  {\tt \{tuvu,miyyer\}@cs.umass.edu} \\}

\date{}

\begin{document}
\maketitle

\input{2019_acl_paragraph_embeddings/sections/abstract}
\input{2019_acl_paragraph_embeddings/sections/introduction}
\input{2019_acl_paragraph_embeddings/sections/probing}
\input{2019_acl_paragraph_embeddings/sections/classification}
\input{2019_acl_paragraph_embeddings/sections/related_work}
\input{2019_acl_paragraph_embeddings/sections/conclusions}
\input{2019_acl_paragraph_embeddings/sections/acknowledgments}

\input{bbl/2019_acl_paragraph_embeddings.bbl}
\appendix
\section{Appendices}
\label{sec:appendix}
\input{2019_acl_paragraph_embeddings/sections/appendix}


\end{document}

%% file: 2019_acl_paragraph_embeddings/sections/abstract.tex
\begin{abstract}
\label{sec:abstract}
While paragraph embedding models are remarkably effective for downstream classification tasks, what they learn and encode into a single vector remains opaque. In this paper, we investigate a state-of-the-art paragraph embedding method proposed by ~\newcite{Zhang:17} and discover that it cannot reliably tell whether a given sentence occurs in the input paragraph or not. We formulate a \emph{sentence content} task to probe for this basic linguistic property and find that even a much simpler bag-of-words method has no trouble solving it. This result motivates us to replace the reconstruction-based objective of~\newcite{Zhang:17} with our sentence content probe objective in a semi-supervised setting. Despite its simplicity, our objective improves over paragraph reconstruction in terms of (1) downstream classification accuracies on benchmark datasets, (2) faster training, and (3) better generalization ability.\footnote{Source code and data are available at \href{https://github.com/tuvuumass/SCoPE}{https://github.com/\\tuvuumass/SCoPE}.}
\end{abstract}

%% file: 2019_acl_paragraph_embeddings/sections/introduction.tex
\section{Introduction}
\label{sec:introduction}
Methods that embed a paragraph into a single vector have been successfully integrated into many NLP applications, including text classification~\citep{Zhang:17}, document retrieval~\citep{Le:14}, and semantic similarity and relatedness~\citep{Dai:15b,Chen:17}. However, downstream performance provides little insight into the kinds of linguistic properties that are encoded by these embeddings. Inspired by the growing body of work on sentence-level linguistic \textit{probe tasks}~\citep{Adi:17,Conneau:18}, we set out to evaluate a state-of-the-art paragraph embedding method using a probe task to measure how well it encodes the identity of the sentences within a paragraph. We discover that the method falls short of capturing this basic property, and that implementing a simple objective to fix this issue improves classification performance, training speed, and generalization ability.

We specifically investigate the paragraph embedding method of ~\newcite{Zhang:17}, which consists of a CNN-based encoder-decoder model~\citep{Sutskever:14} paired with a reconstruction objective to learn powerful paragraph embeddings that are capable of accurately reconstructing long paragraphs. This model significantly improves downstream classification accuracies, outperforming LSTM-based alternatives~\citep{Li:15}.

How well do these embeddings encode \textit{whether or not a given sentence appears in the paragraph}?
\newcite{Conneau:18} show that such identity information is correlated with performance on downstream sentence-level tasks. We thus design a probe task to measure the extent to which this \emph{ sentence content} property is captured in a paragraph embedding. Surprisingly, our experiments (Section~\ref{sec:probing}) reveal that despite its impressive downstream performance, the model of \newcite{Zhang:17} substantially underperforms a simple bag-of-words model on our sentence content probe.

Given this result, it is natural to wonder whether the sentence content property is actually useful for downstream classification. To explore this question, we move to a semi-supervised setting by pre-training the paragraph encoder in Zhang et al.'s model \shortcite{Zhang:17} on either our sentence content objective or its original reconstruction objective, and then optionally fine-tuning it on supervised classification tasks (Section~\ref{sec:classification}). Sentence content significantly improves over reconstruction on standard benchmark datasets both with and without fine-tuning; additionally, this objective is four times faster to train than the reconstruction-based variant. Furthermore, pre-training with sentence content substantially boosts generalization ability: fine-tuning a pre-trained model on just 500 labeled reviews from the Yelp sentiment dataset surpasses the accuracy of a purely supervised model trained on 100,000 labeled reviews.

Our results indicate that incorporating probe objectives into downstream models might help improve both accuracy and efficiency, which we hope will spur more linguistically-informed research into paragraph embedding methods.

%% file: 2019_acl_paragraph_embeddings/sections/probing.tex
\section{Probing paragraph embeddings for sentence content}
\label{sec:probing}
In this section, we first fully specify our probe task before comparing the model of~\newcite{Zhang:17} to a simple bag-of-words model. Somewhat surprisingly, the latter substantially outperforms the former despite its relative simplicity.
\vspace{-0.5mm}
\subsection{Probe task design}
Our proposed \emph{sentence content} task is a paragraph-level analogue to the word content task of~\newcite{Adi:17}: given embeddings\footnote{computed using the same embedding method} $\bvec{p}, \bvec{s}$ of a paragraph $p$ and a candidate sentence $s$, respectively, we train a classifier to predict whether or not $s$ occurs in $p$. Specifically, we construct a binary classification task in which the input is $[\bvec{p};\bvec{s}]$, the concatenation of $\bvec{p}$ and $\bvec{s}$. This task is balanced: for each paragraph $p$ in our corpus, we create one positive instance by sampling a sentence $s^+$ from $p$ and one negative instance by randomly sampling a sentence $s^-$ from another paragraph $p'$. As we do not perform any fine-tuning of the base embedding model, our methodology is agnostic to the choice of model.
\vspace{-0.5mm}
\subsection{Paragraph embedding models}
Armed with our probe task, we investigate the following embedding methods.\footnote{We experiment with several other models in  Appendix~\ref{sec:appendix1}, including an LSTM-based encoder-decoder model, a variant of Paragraph Vector~\citep{Le:14}, and BOW models using pre-trained word representations.}
\vspace{-0.5mm}
\paragraph{\newcite{Zhang:17} (CNN-R):} This model uses  a multi-layer convolutional encoder to compute a single vector embedding $\bvec{p}$ of an input paragraph $p$ and a multi-layer deconvolutional decoder that mirrors the convolutional steps in the encoding stage to reconstruct the tokens of $p$ from $\bvec{p}$. We refer readers to Zhang et al. \shortcite{Zhang:17} for a detailed description of the model architecture. For a more intuitive comparison in our experiments, we denote this model further as CNN-R instead of CNN-DCNN as in the original paper. In all experiments, we use their publicly available code.\footnote{\href{https://github.com/dreasysnail/textCNN_public}{https://github.com/dreasysnail/textCNN\_public}}
\vspace{-1mm}
\paragraph{Bag-of-words (BoW):} This model is simply an average of the word vectors learned by a trained CNN-R model. BoW models have been shown to be surprisingly good at sentence-level probe tasks \citep{Adi:17,Conneau:18}.
\vspace{-1mm}
\subsection{Probe experimental details}
Paragraphs to train our classifiers are extracted from the Hotel Reviews corpus~\citep{Li:15}, which has previously been used for evaluating the quality of paragraph embeddings \cite{Li:15,Zhang:17}. We only consider paragraphs that have at least two sentences. Our dataset has 346,033 training paragraphs, 19,368 for validation, and 19,350 for testing. The average numbers of sentences per paragraph, tokens per paragraph, and tokens per sentence are 8.0, 123.9, and 15.6, respectively. The vocabulary contains 25,000 tokens. To examine the effect of the embedding dimensionality $d$ on the results, we trained models with $d \in \{100, 300, 500, 700, 900\}$.

Each classifier is a feed-forward neural network with a single 300-$d$ ReLu layer. We use a mini-batch size of 32, Adam optimizer \cite{Kingma:15} with a learning rate of 2$e$-4, and a dropout rate of 0.5 \cite{Srivastava:14}. We trained classifiers for a maximum of 100 epochs with early stopping based on validation performance.

\subsection{BoW outperforms CNN-R on sentence content}
\input{2019_acl_paragraph_embeddings/figs/fig1tex}
\input{2019_acl_paragraph_embeddings/figs/fig2tex}
Our probe task results are displayed in Figure~\ref{fig1}. Interestingly, BoW performs significantly better than CNN-R,
achieving an accuracy of 87.2\% at 900 dimensions, compared to only 66.4\% for CNN-R. We hypothesize that much of BoW's success is because it is easier for the model to perform approximate string matches between the candidate sentence and text segments within the paragraph than it is for the highly non-linear representations of CNN-R.

To investigate this further, we repeat the experiment, but exclude the sentence $s^+$ from the paragraph $p$ during both training and testing. As we would expect (see Table~\ref{tbl1}), BoW's performance degrades significantly (20.6\% absolute) with $s^+$ excluded from $p$, whereas CNN-R experiences a more modest drop (3.6\%). While BoW still outperforms CNN-R in this new setting, the dramatic drop in accuracy suggests that it relies much more heavily on low-level matching.
\input{2019_acl_paragraph_embeddings/tbls/tbl1}

%% file: 2019_acl_paragraph_embeddings/figs/fig1tex.tex
\begin{figure}[t]
\centering
\includegraphics[width=0.44\textwidth]{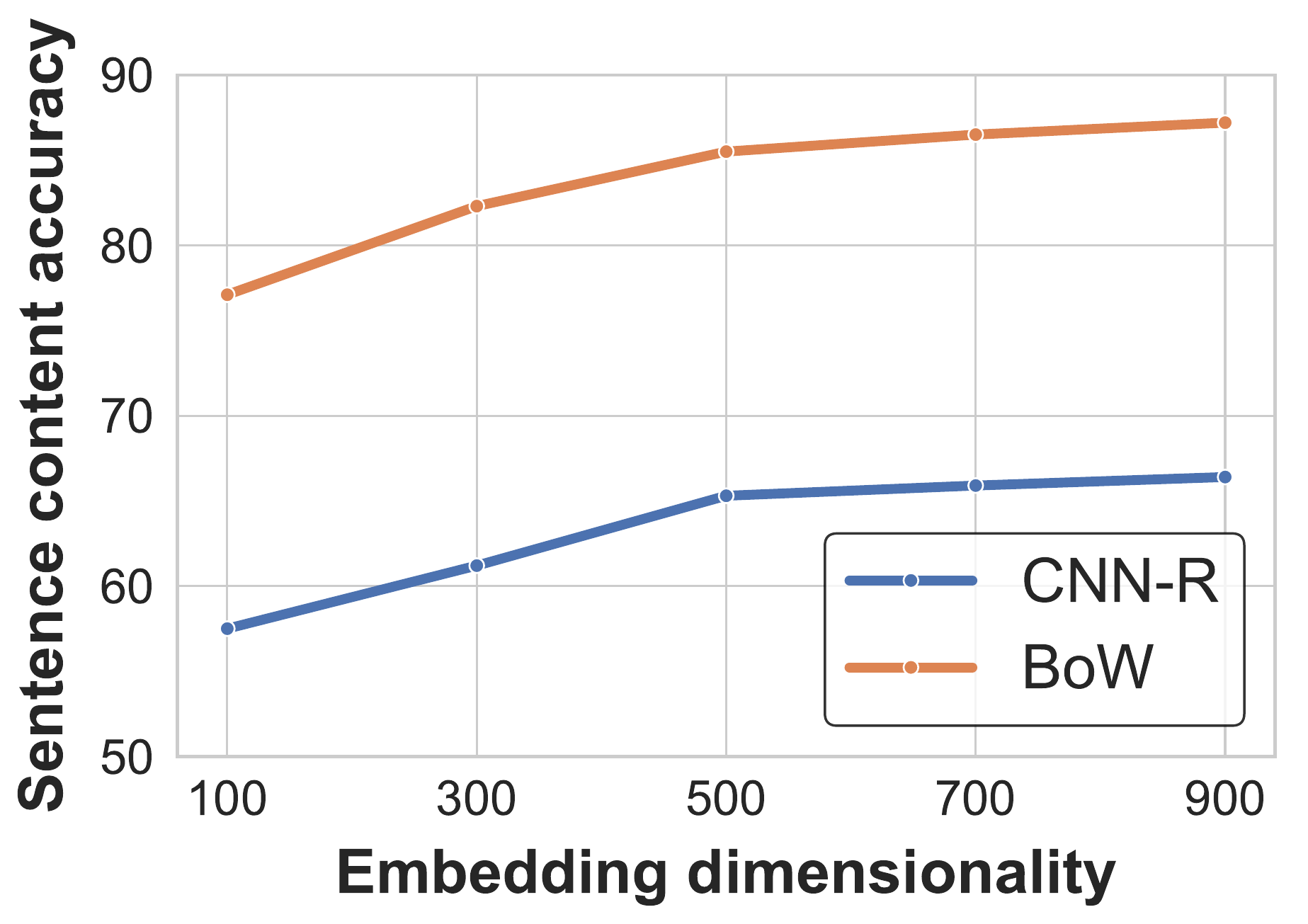}
\caption{Probe task accuracies across representation dimensions. BoW surprisingly outperforms the more complex model CNN-R.}
\label{fig1}
\end{figure}

%% file: 2019_acl_paragraph_embeddings/figs/fig2tex.tex
\begin{figure*}[t]
\centering
\includegraphics[width=0.65\textwidth]{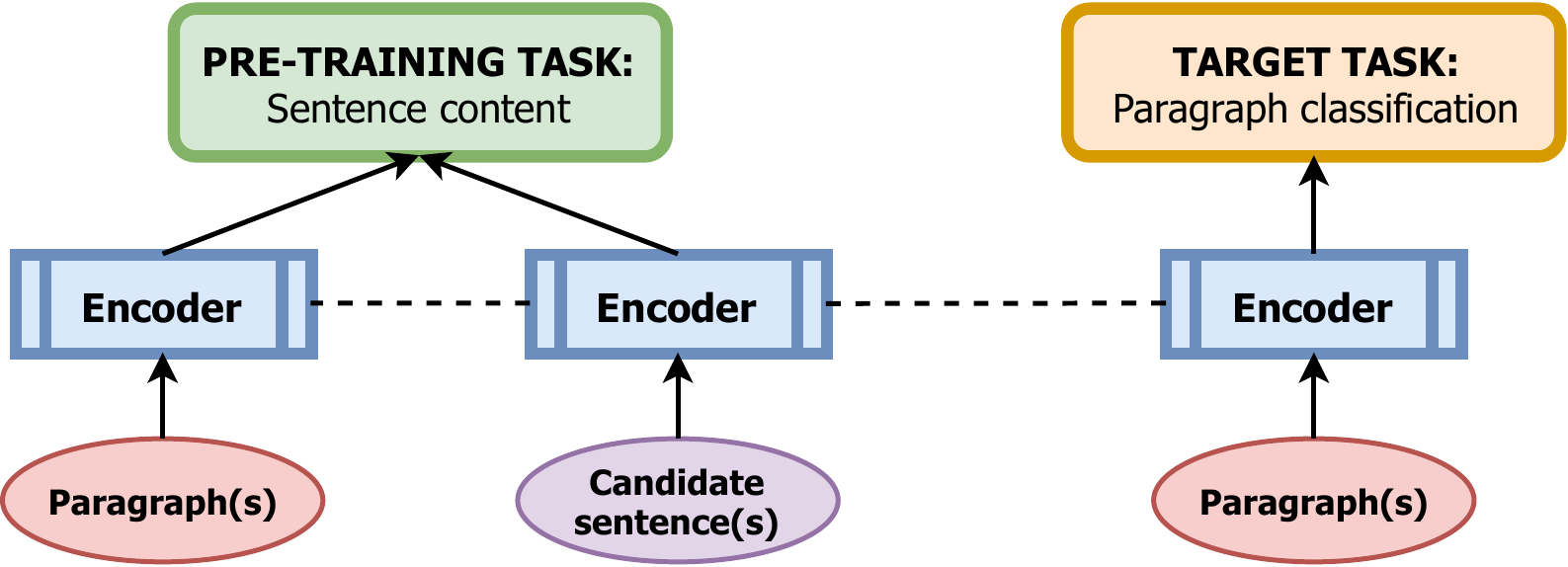}
\caption{A visualization of our semi-supervised approach. We first train the CNN encoder (shown as two copies with shared parameters) on unlabeled data using our sentence content objective. The encoder is then used for downstream classification tasks.}
\label{fig2}
\end{figure*}

%% file: 2019_acl_paragraph_embeddings/tbls/tbl1.tex
\begin{table}[t]
\centering
\begin{adjustbox}{max width=0.44\textwidth}
 \begin{tabular}{ c c c }
  \toprule
Setting & CNN-R & BoW \\ [0.5ex] 
 \midrule
Without $s^{+}$ excluded from $p$ & 61.2 & \textbf{82.3} \\
With $s^{+}$ excluded from $p$ & 57.5 & \textbf{61.7} \\
\bottomrule
 \end{tabular}
 \end{adjustbox}
  \caption{Probe task accuracies without and with $s^{+}$ excluded from $p$, measured at $d=300$. BoW's accuracy degrades quickly in the latter case, suggesting that it relies much more on low-level matching.}
  \label{tbl1}
\end{table}

%% file: 2019_acl_paragraph_embeddings/sections/classification.tex
\section{Sentence content improves paragraph classification}
\label{sec:classification}
Motivated by our probe results, we further investigate whether incorporating the sentence content property into a paragraph encoder can help increase downstream classification accuracies. We propose a semi-supervised approach by pre-training the encoder of CNN-R using our sentence content objective, and optionally fine-tuning it on different classification tasks. A visualization of this procedure can be seen in Figure~\ref{fig2}. We compare our approach (henceforth \textbf{CNN-SC}) without and with fine-tuning against CNN-R, which uses a reconstruction-based objective.\footnote{Here, we use unsupervised pre-training as it allows us to isolate the effects of the unsupervised training objectives. \newcite{Zhang:17} implemented auxiliary unsupervised training as an alternative form of semi-supervised learning. We tried both strategies and found that they performed similarly.} We report comparisons on three standard paragraph classification datasets: Yelp Review Polarity (Yelp), DBPedia, and Yahoo! Answers (Yahoo)~\citep{Zhang:15}, which are instances of common classification tasks, including sentiment analysis and topic classification. Table~\ref{tbl2} shows the statistics for each dataset. Paragraphs from each training set without labels were used to generate training data for unsupervised pre-training.
\input{2019_acl_paragraph_embeddings/tbls/tbl2}
\paragraph{Sentence content significantly improves over reconstruction on both in-domain and out-of-domain data}
\input{2019_acl_paragraph_embeddings/figs/fig3tex}
We first investigate how useful each pre-training objective is for downstream classification without any fine-tuning by simply training a classifier on top of the frozen pre-trained CNN encoder. We report the best downstream performance for each model across different numbers of pre-training epochs. The first row of Table~\ref{tbl3} shows the downstream accuracy on Yelp when the whole unlabeled data of the Yelp training set is used for unsupervised pre-training. Strikingly, CNN-SC achieves an accuracy of 90.0\%, outperforming CNN-R by a large margin. Additionally, sentence content is four times as fast to train as the computationally-expensive reconstruction objective.\footnote{This objective requires computing a probability distribution over the whole vocabulary for every token of the paragraph, making it prohibitively slow to train.} 
\input{2019_acl_paragraph_embeddings/tbls/tbl3}
Are representations obtained using these objectives more useful when learned from in-domain data? To examine the dataset effect, we repeat our experiments using paragraph embeddings pre-trained using these objectives on a subset of Wikipedia (560K paragraphs). The second row of Table~\ref{tbl3} shows that both approaches suffer a drop in downstream accuracy when pre-trained on out-of-domain data. Interestingly, CNN-SC still performs best, indicating that sentence content is more suitable for downstream classification.

Another advantage of our sentence content objective over reconstruction is that it better correlates to downstream accuracy (see Appendix~\ref{sec:appendix2}). For reconstruction, there is no apparent correlation between BLEU and downstream accuracy;  while BLEU increases with the number of epochs, the downstream performance quickly begins to decrease. This result indicates that early stopping based on BLEU is not feasible with reconstruction-based pre-training objectives.
\paragraph{With fine-tuning, CNN-SC substantially boosts accuracy and generalization}
\input{2019_acl_paragraph_embeddings/tbls/tbl4}
We switch gears now to our fine-tuning experiments. Specifically, we take the CNN encoder pre-trained using our sentence content objective and then fine-tune it on downstream classification tasks with supervised labels. While our previous version of CNN-SC created just a single positive/negative pair of examples from a single paragraph, for our fine-tuning experiments we create a pair of examples from \emph{every sentence} in the paragraph to maximize the training data. For each task, we compare against the original CNN-R model in~\citep{Zhang:17}. Figure~\ref{fig3} shows the model performance with fine-tuning on 0.1\% to 100\% of the training set of each dataset. One interesting result is that CNN-SC relies on very few training examples to achieve comparable accuracy to the purely supervised CNN model. For instance, fine-tuning CNN-SC using just 500 labeled training examples surpasses the accuracy of training from scratch on 100,000 labeled examples, indicating that the sentence content encoder generalizes well. CNN-SC also outperforms CNN-R by large margins when only small amounts of labeled training data are available. Finally, when all labeled training data is used, CNN-SC  achieves higher classification accuracy than CNN-R on all three datasets (Table~\ref{tbl4}).

While CNN-SC exhibits a clear preference for target task unlabeled data (see Table~\ref{tbl3}), we can additionally leverage large amounts of unlabeled general-domain data by incorporating pre-trained word representations from language models into CNN-SC. Our results show that further improvements can be achieved by training the sentence content objective on top of the pre-trained language model representations from ULMFiT~\cite{Howard:18} (see Appendix~\ref{sec:appendix3}), indicating that our sentence content objective learns complementary information. On Yelp, it exceeds the performance of training from scratch on the whole labeled data (560K examples) with only 0.1\% of the labeled data.

\paragraph{CNN-SC implicitly learns to distinguish between class labels}
The substantial difference in downstream accuracy between pre-training on in-domain and out-of-domain data (Table~\ref{tbl3}) implies that the sentence content objective is implicitly learning to distinguish between class labels (e.g., that a candidate sentence with negative sentiment is unlikely to belong to a paragraph with positive sentiment). If true, this result implies that CNN-SC prefers not only in-domain data but also a representative sample of paragraphs from all class labels. To investigate, we conduct an additional experiment that restricts the class label from which negative sentence candidates $s^{-}$ are sampled. We experiment with two sources of $s^{-}$: (1) paragraphs of the same class label as the probe paragraph (CNN-SC$^{-}$), and  (2) paragraphs from a different class label (CNN-SC$^{+}$). Figure~\ref{fig4} reveals that the performance of CNN-SC drops dramatically when trained on the first dataset and improves when trained on the second dataset, which confirms our hypothesis.
\input{2019_acl_paragraph_embeddings/figs/fig4tex}

%% file: 2019_acl_paragraph_embeddings/tbls/tbl2.tex
\begin{table}[t!]
\centering
\begin{adjustbox}{max width=0.44\textwidth}
 \begin{tabular}{ c c c c}
  \toprule

Dataset  & Type & \# classes & \# examples\\ [0.5ex] 
 \midrule
Yelp & Sentiment & 2 & 560K \\
DBpedia & Topic & 14 & 560K \\
Yahoo & Topic & 10 & 1.4M \\

\bottomrule
 \end{tabular}
 \end{adjustbox}
  \caption{Properties of the text classification datasets used for our evaluations.}
  \label{tbl2}
\end{table}

%% file: 2019_acl_paragraph_embeddings/figs/fig3tex.tex
\begin{figure*}[t]
\centering
\begin{subfigure}[t]{0.32\textwidth}
        \includegraphics[width=\linewidth]{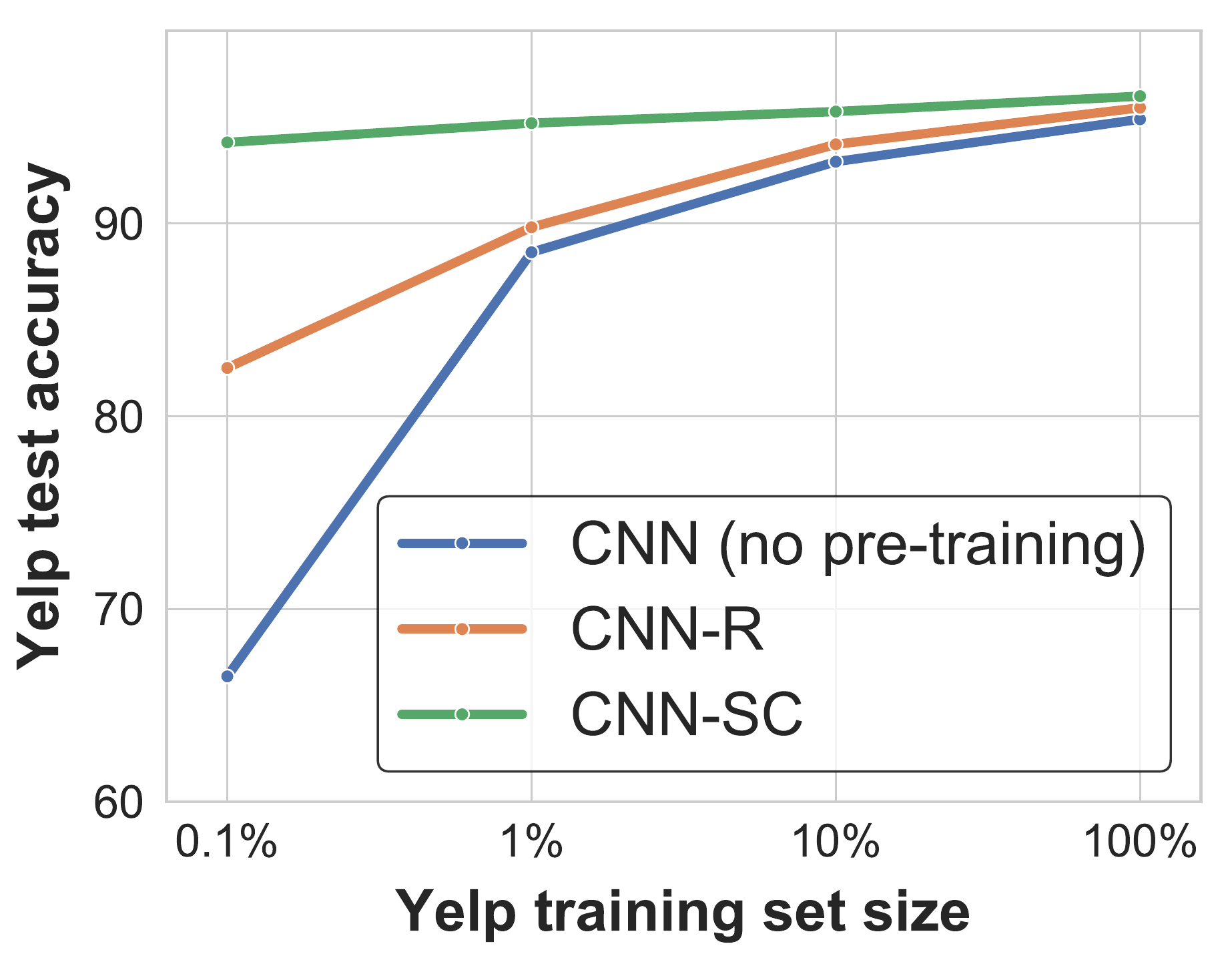}
    \end{subfigure}%
    ~
    \begin{subfigure}[t]{0.32\textwidth}
        \includegraphics[width=\linewidth]{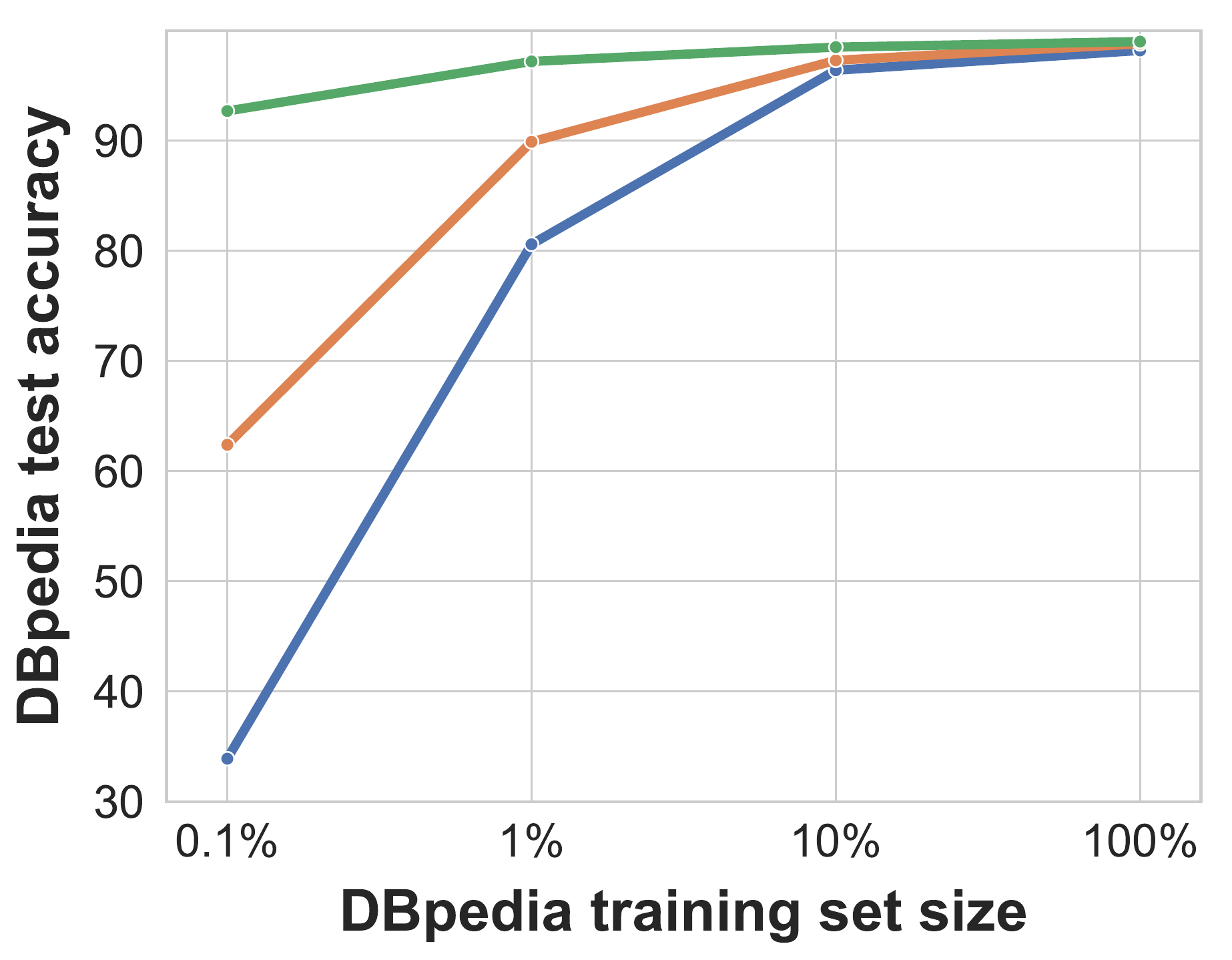}
  \end{subfigure}
   \begin{subfigure}[t]{0.32\textwidth}
        \includegraphics[width=\linewidth]{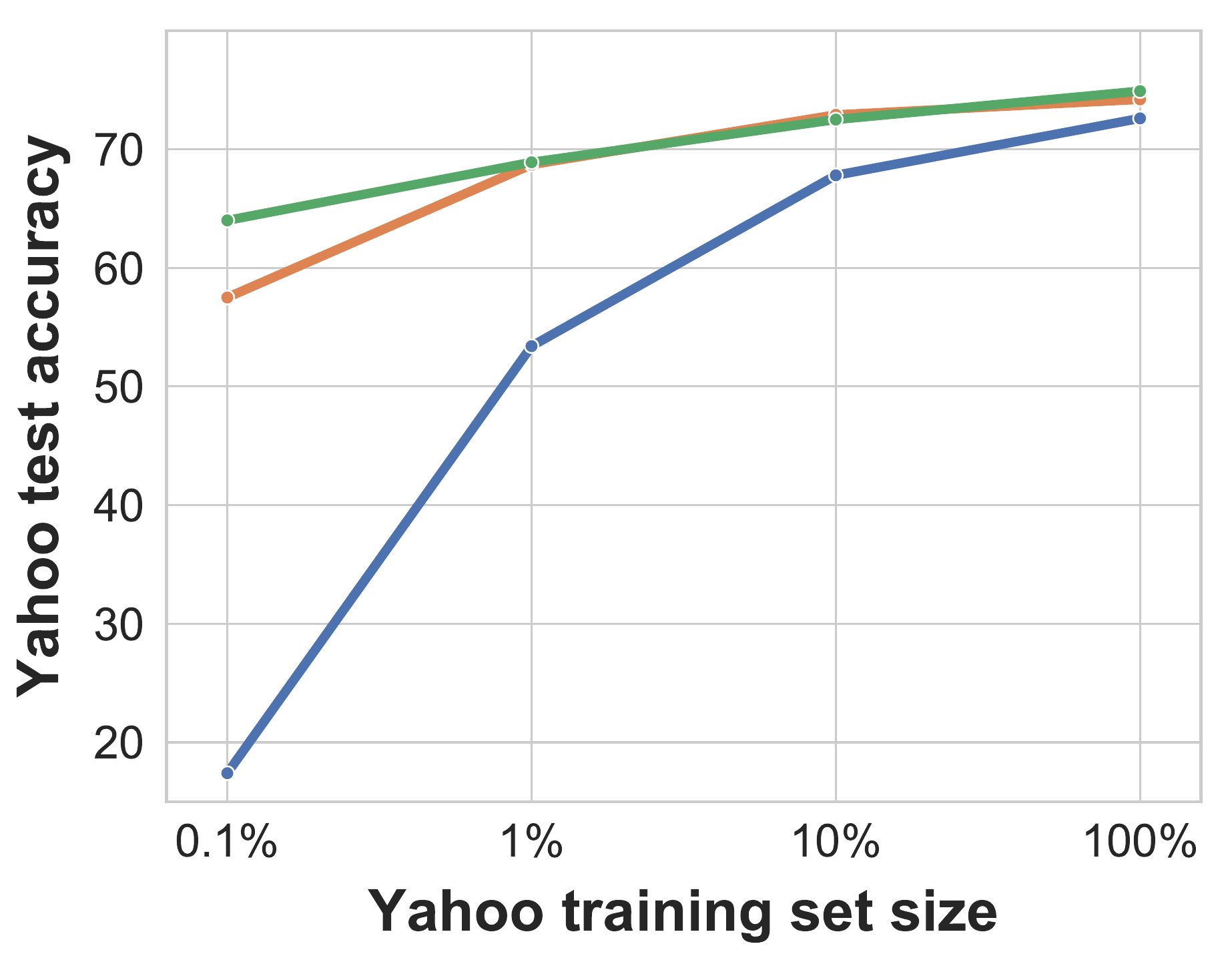}
    \end{subfigure}   
\caption{CNN-SC substantially improves generalization ability. Results of CNN-R are taken from~\citet{Zhang:17}.}
\label{fig3}
\vspace*{-2mm}
\end{figure*}

%% file: 2019_acl_paragraph_embeddings/tbls/tbl3.tex
\begin{table}[t!]
\centering
\begin{adjustbox}{max width=0.44\textwidth}
 \begin{tabular}{ c c c }
  \toprule

Pre-training  & CNN-R & CNN-SC\\ [0.5ex] 
 \midrule
On Yelp & 67.4 & \textbf{90.0}\\
On Wikipedia & 61.4 & \textbf{65.7}\\
Wall-clock speedup & 1x & \textbf{4x}\\ 

\bottomrule
 \end{tabular}
 \end{adjustbox}
  \caption{Yelp test accuracy (without fine-tuning). CNN-SC significantly improves over CNN-R.}
  \label{tbl3}
\end{table}

%% file: 2019_acl_paragraph_embeddings/tbls/tbl4.tex
\begin{table}[t!]
\centering
\begin{adjustbox}{max width=0.48\textwidth}
 \begin{tabular}{l c c c }
 \toprule
Model & Yelp & DBPedia &  Yahoo \\ [0.5ex] 
\midrule
\multicolumn{4}{c}{\emph{purely supervised w/o external data}} \\
ngrams TFIDF	&95.4	&98.7	&68.5\\
Large Word ConvNet	&95.1	&98.3	&70.9\\
Small Word ConvNet	&94.5	&98.2	&70.0\\
Large Char ConvNet	 &94.1	&98.3	&70.5\\
Small Char ConvNet	&93.5	&98.0	&70.2\\
SA-LSTM (word level)	&NA	&98.6	&NA\\
Deep ConvNet	&95.7	&98.7	&73.4\\
CNN \cite{Zhang:17}	&95.4 &	98.2	&72.6\\
\midrule
\multicolumn{4}{c}{\emph{pre-training + fine-tuning w/o external data}} \\
CNN-R \cite{Zhang:17}		&96.0	&98.8	&74.2\\
CNN-SC (ours) & \textbf{96.6}	& \textbf{99.0}	& \textbf{74.9}\\
\midrule
\multicolumn{4}{c}{\emph{pre-training + fine-tuning w/ external data}} \\
ULMFiT \cite{Howard:18}	 & 97.8 & 99.2 & NA \\
\bottomrule
 \end{tabular}
\end{adjustbox}
   \caption{CNN-SC outperforms other baseline models that do not use external data, including CNN-R. All baseline models are taken from~\citet{Zhang:17}.}
  \label{tbl4}
 \vspace*{-2mm}
\end{table}

%% file: 2019_acl_paragraph_embeddings/figs/fig4tex.tex
\begin{figure}[t]
\centering
\includegraphics[width=0.44\textwidth]{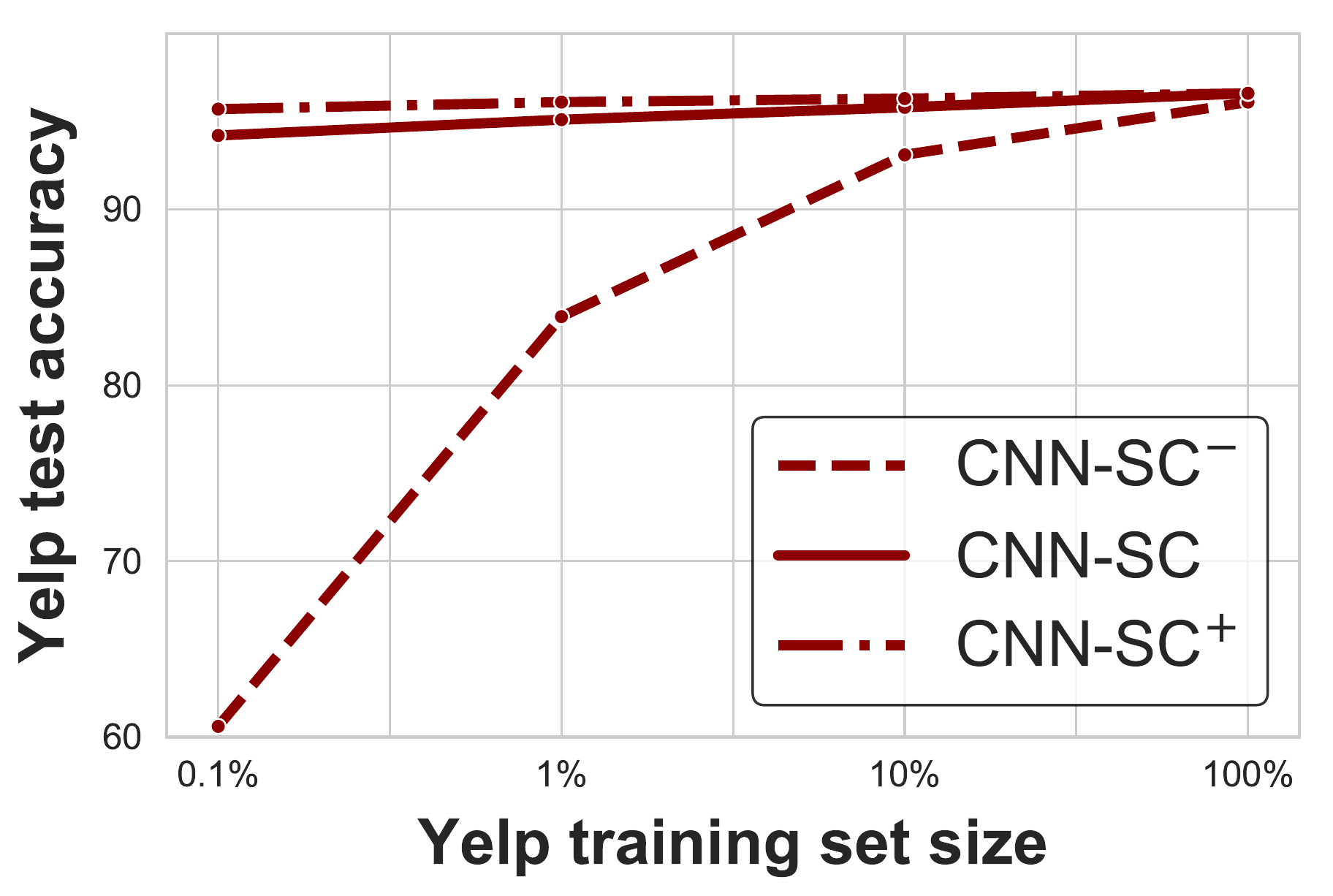}
\caption{CNN-SC implicitly learns to distinguish between class labels.}
\label{fig4}
\vspace*{-2mm}
\end{figure}

%% file: 2019_acl_paragraph_embeddings/sections/related_work.tex
\section{Related work}
\label{sec:related_work}
\paragraph{Text embeddings and probe tasks}
A variety of methods exist for obtaining fixed-length dense vector representations of words~\cite[e.g.,][]{Mikolov:13,Pennington:14,Peters:18}, sentences~\cite[e.g.,][]{Kiros:15,Conneau:17,Subramanian:18,Cer:18}, and larger bodies of text~\cite[e.g.,][]{Le:14,Dai:15b,Iyyer:15,Li:15,Chen:17,Zhang:17} that significantly improve various downstream tasks. To analyze word and sentence embeddings, recent work has studied classification tasks that probe them for various linguistic properties~\cite{Shi:16,Adi:17,Belinkov:17a,Belinkov:17b,Conneau:18,Tenney:19}. In this paper, we extend the notion of probe tasks to the paragraph level.
\paragraph{Transfer learning}
Another line of related work is transfer learning, which has been the driver of recent successes in NLP. Recently-proposed objectives for transfer learning include surrounding sentence prediction ~\cite{Kiros:15}, paraphrasing~\cite{Wieting:17}, entailment~\cite{Conneau:17}, machine translation~\cite{McCann:17}, discourse~\cite{Jernite:17,Nie:17}, and language modeling~\cite{Peters:18,Radford:18,Devlin:18}.

%% file: 2019_acl_paragraph_embeddings/sections/conclusions.tex
\section{Conclusions and Future work}
\label{sec:conclusions}
In this paper, we evaluate a state-of-the-art paragraph embedding model, based on how well it captures the sentence identity within a paragraph. Our results indicate that the model is not fully aware of this basic property, and that implementing a simple objective to fix this issue improves classification performance, training speed, and generalization ability. Future work can investigate other embedding methods with a richer set of probe tasks, or explore a wider range of downstream tasks.

%% file: 2019_acl_paragraph_embeddings/sections/acknowledgments.tex
\section*{Acknowledgments}

We thank the anonymous reviewers, Kalpesh Krishna, Nader Akoury, and the members of the UMass NLP reading group for their helpful comments.

%% file: 2019_acl_paragraph_embeddings/sections/appendix.tex
\subsection{BoW models outperform more complex models on our sentence content probe}
\label{sec:appendix1}
In addition to the paragraph embedding models presented in the main paper, we also experiment with the following embedding methods:
\paragraph{LSTM-R:} We consider an LSTM~\cite{Hochreiter:97} encoder-decoder model paired with a reconstruction objective.  Specifically, we implement a single-layer bidirectional LSTM encoder and a two-layer unidirectional LSTM decoder. Paragraph representations are computed from the encoder's final hidden state.
\paragraph{Doc2VecC:} This model~\cite{Chen:17} represents a document as an average of randomly-sampled words from within the document. The method introduces a corruption mechanism that favors rare but important words while suppressing frequent but uninformative ones. Doc2VecC was found to outperform other unsupervised BoW-style algorithms, including Paragraph Vector~\cite{Le:14}, on downstream tasks.
\paragraph{Other BoW models:} We also consider other BoW models with pre-trained word embeddings or contextualized word representations, including Word2Vec~\cite{Mikolov:13}, Glove~\cite{Pennington:14}, and ELMo~\cite{Peters:18}. Paragraph embeddings are computed as the average of the word vectors. For ELMo, we take the average of the layers.

The results of our sentence content probe task are summarized in Table~\ref{tbl5}.
\input{2019_acl_paragraph_embeddings/tbls/tbl5}
\subsection{Sentence content better correlates to downstream accuracy than reconstruction}
\label{sec:appendix2}
See Figure~\ref{fig5}.
\input{2019_acl_paragraph_embeddings/figs/fig5tex}
\subsection{Further improvements by training sentence content on top of pre-trained language model representations}
\label{sec:appendix3}
\input{2019_acl_paragraph_embeddings/figs/fig6tex}
Figure \ref{fig6} shows that further improvements can be achieved by training sentence content on top of the pre-trained language model representations from ULMFiT~\cite{Howard:18} on Yelp and IMDB~\citep{Maas:11} datasets, indicating that our sentence content objective learns complementary information.\footnote{Here, we do not perform target task classifier fine-tuning to isolate the effects of our sentence content objective.} On Yelp, it exceeds the performance of training from scratch on the whole labeled data (560K examples) with only 0.1\% of the labeled data.

%% file: 2019_acl_paragraph_embeddings/tbls/tbl5.tex
\begin{table}[t]
\centering
\begin{adjustbox}{max width=0.48\textwidth}
 \begin{tabular}{l c c c}
  \toprule
\textbf{Model} &  \textbf{Dimensionality} & \textbf{Accuracy} \\ [0.5ex]
 \midrule
 Random  & -- & 50.0 \\
 \midrule
 \multicolumn{3}{c}{\emph{trained on paragraphs from Hotel Reviews}}\\
 CNN-R & 900 & 66.4 \\
 BoW (CNN-R) & 900 & 87.2\\
 LSTM-R & 900 & 65.4 \\
 Doc2VecC & 900 & 90.8 \\
 \midrule
\multicolumn{3}{c}{\emph{pre-trained on other datasets}}\\ 
Word2Vec-avg & 300 & 83.2\\
GloVe-avg & 300 & 84.6\\
ELMo-avg & 1024 & 88.1 \\
\bottomrule
 \end{tabular}
 \end{adjustbox}
  \caption{Sentence content accuracy for different paragraph embedding methods. BoW models outperform more complex models.}
  \label{tbl5}
\end{table}

%% file: 2019_acl_paragraph_embeddings/figs/fig5tex.tex
\begin{figure}[h!]
\centering
\includegraphics[width=0.48\textwidth]{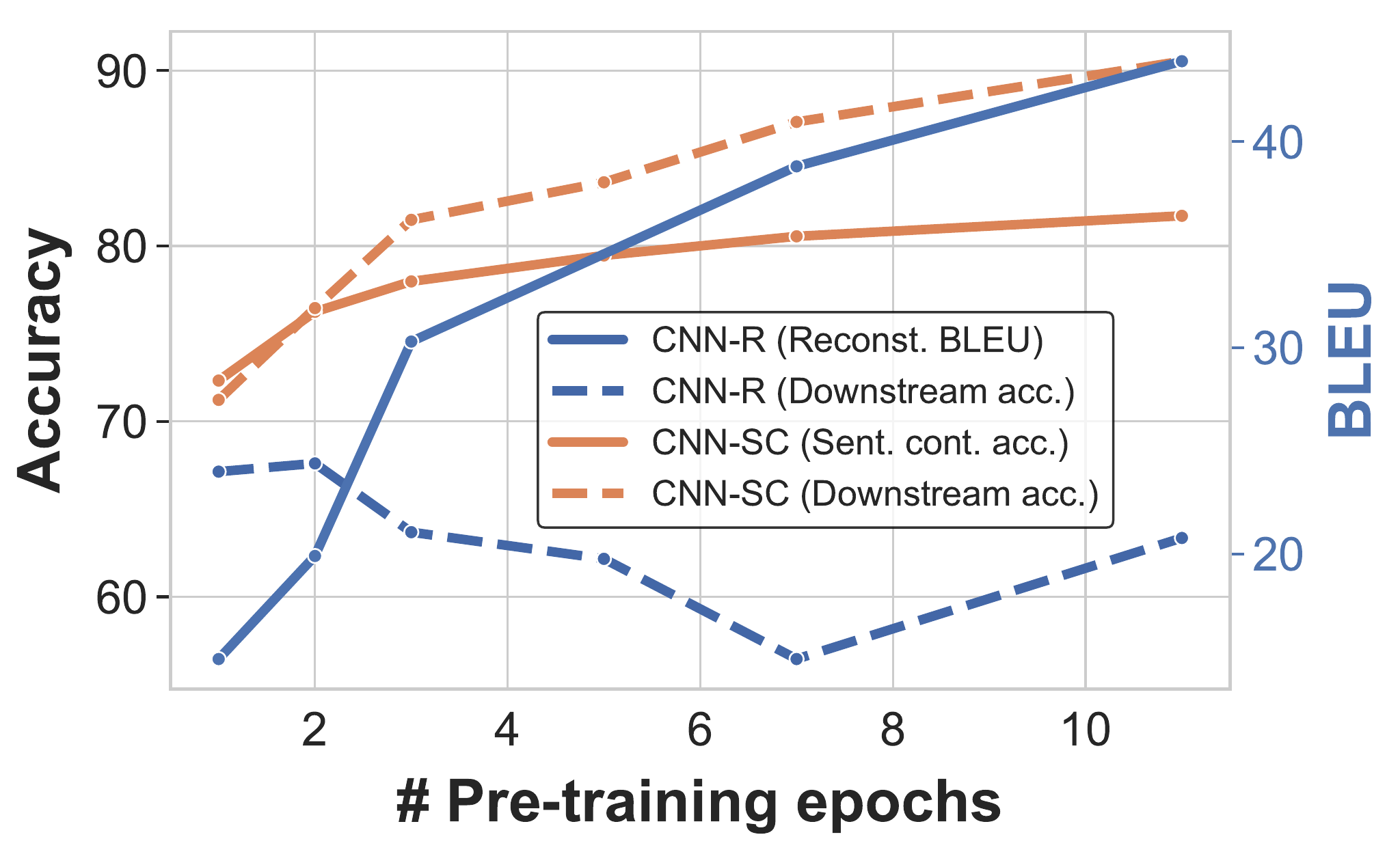}
\caption{Pre-training performance vs. downstream accuracy on Yelp. Performance measured on validation data. There is no apparent correlation between BLEU and downstream accuracy.}
\label{fig5}
\end{figure}

%% file: 2019_acl_paragraph_embeddings/figs/fig6tex.tex
\begin{figure}[t!]
\centering
\begin{subfigure}[t]{0.44\textwidth}
        \includegraphics[width=\linewidth]{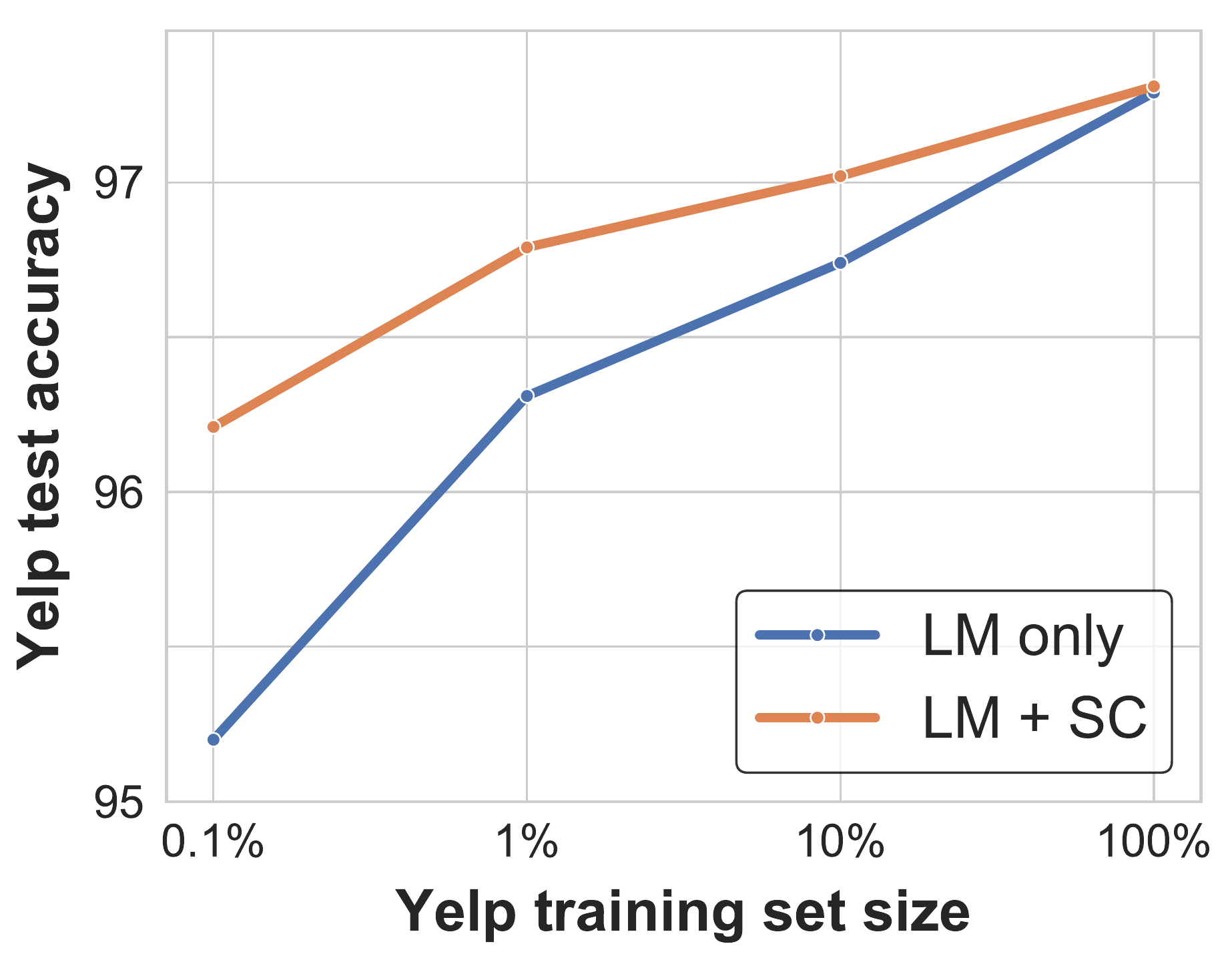}
    \end{subfigure} \\
    \begin{subfigure}[t]{0.44\textwidth}
        \includegraphics[width=\linewidth]{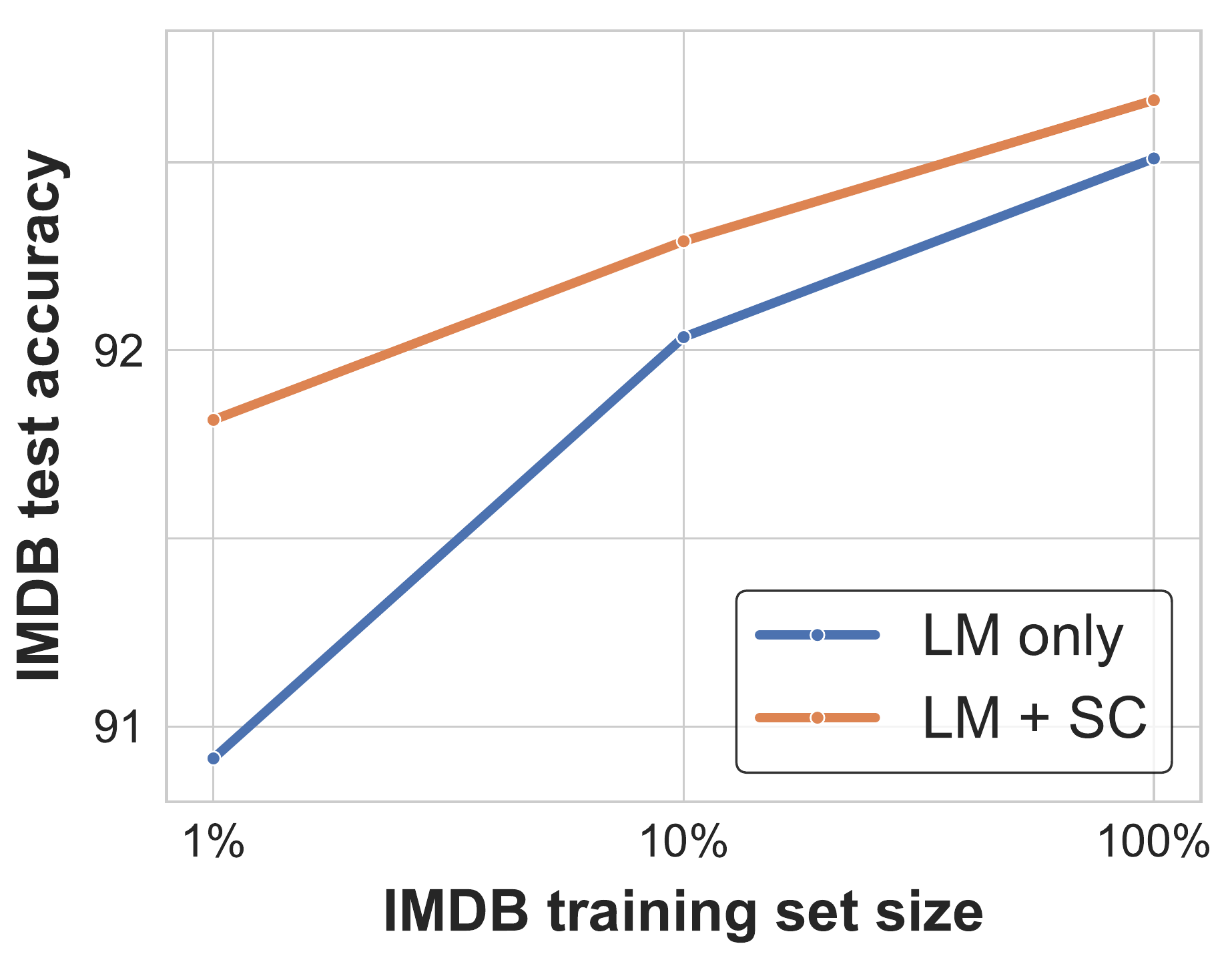}
  \end{subfigure}
\caption{Further improvements can be achieved by training sentence content (SC) on top of the pre-trained language model (LM) representations from ULMFiT~\cite{Howard:18}.}
\label{fig6}
\end{figure}